\def\year{2020}\relax
\documentclass[letterpaper]{article} 
\usepackage{aaai20}  
\usepackage{times}  
\usepackage{helvet} 
\usepackage{courier}  
\usepackage[hyphens]{url}  
\usepackage{graphicx} 
\usepackage{graphicx}  
\usepackage{amsmath,amsthm,amssymb}
\usepackage[dvipsnames]{xcolor}
\newcommand\V{{\bf V}}
\newcommand\Z{{\bf Z}}
\newcommand\X{{\bf X}}
\newcommand\y{{\bf y}}
\newcommand\z{{\bf z}}
\newcommand\Y{{\bf Y}}
\newcommand\x{{\bf x}}
\newcommand\bv{{\bf v}}
\newcommand\bmu{{\bf \mu}}
\newcommand\argmin{\text{argmin}}
\title{Application of Deep Learning-based Interpolation Methods to Nearshore Bathymetry}
\author{Yizhou Qian,\textsuperscript{\rm 1} 
Mojtaba Forghani,\textsuperscript{\rm 1} 
Jonghyun Lee\textsuperscript{2} \\ 
\Large 
\textbf{Matthew Farthing,\textsuperscript{\rm 3} 
Tyler Hesser,\textsuperscript{\rm 3} 
Peter K.\ Kitanidis,\textsuperscript{\rm 1} 
Eric F.\ Darve\textsuperscript{\rm 1}}  \\ 
\textsuperscript{\rm 1}Stanford University,
\textsuperscript{\rm 2}University of Hawaii at Manoa,
\textsuperscript{\rm 3}US Army Engineer Research and Development Center\\
\textsuperscript{\rm 1}\{yzqian, mojtaba, darve, peterk\}@stanford.edu,
\textsuperscript{\rm 2}jonghyun.harry.lee@hawaii.edu \\
\textsuperscript{\rm 3}\{matthew.w.farthing, tyler.hesser\}@erdc.dren.mil
}

\begin{document}
\setcounter{secnumdepth}{3}
\maketitle

\begin{abstract}
Nearshore bathymetry, the topography of the ocean floor in coastal zones, is vital for predicting the surf zone hydrodynamics and for route planning to avoid subsurface features. Hence, it is increasingly important for a wide variety of applications, including shipping operations, coastal management, and risk assessment. However, direct high resolution surveys of nearshore bathymetry are rarely performed due to budget constraints and logistical restrictions. Another option when only sparse observations are available is to use Gaussian Process regression (GPR), also called Kriging. But GPR has difficulties recognizing patterns with sharp gradients, like those found around sand bars and submerged objects, especially when observations are sparse. In this work, we present several deep learning-based techniques to estimate nearshore bathymetry with sparse, multi-scale measurements. We propose a Deep Neural Network (DNN) to compute posterior estimates of the nearshore bathymetry, as well as a conditional Generative Adversarial Network (cGAN) that samples from the posterior distribution. We train our neural networks based on synthetic data generated from nearshore surveys provided by the U.S.\ Army Corps of Engineer Field Research Facility (FRF) in Duck, North Carolina. We compare our methods with Kriging on real surveys as well as surveys with artificially added sharp gradients. Results show that direct estimation by DNN gives better predictions than Kriging in this application. We use bootstrapping with DNN for uncertainty quantification. We also propose a method, named DNN-Kriging, that combines deep learning with Kriging and shows further improvement of the posterior estimates.
\end{abstract}

\section{Introduction}
Nearshore bathymetry, or the topography of ocean floor in coastal zones, is critical in many areas, including geomorphology \cite{finkl2005}, harbor managements \cite{grifoll2011}, and flood risk assessment \cite{casas2006}. Accurate estimations of nearshore bathymetry are increasingly important due to the expansion of coastal activities and the improvement of sensor technologies. However, due to budget and logistical limitations, direct high-resolution surveys of bathymetry by using multi-beam sonar equipment \cite{casas2006} are rarely performed. Thus, indirect techniques are used to predict nearshore bathymetry, such as spatial interpolation \cite{merwade2009effect,curtarelli2015assessment} and remote sensing based methods \cite{misra2003approach,gholamalifard2013remotely}.

For remote-sensing based methods, several approaches have been proposed, including airborne bathymetric LiDAR systems \cite{mckean2009remote}, drifting buoys \cite{emery2010autonomous}, and satellite remote sensing imagery \cite{pacheco2015retrieval,pereira2019estimation}. These methods have the advantage of providing an extensive coverage of the area of interest. Yet, application of these methods is limited by environmental or climatic constraints, such as water turbidity and meteorological conditions, and strict requirements for expert personnel, equipment, and maintenance.

For spatial interpolation methods, one of the most widely used stochastic techniques is Kriging, or Gaussian Process Regression \cite{camps2016survey}. \citeauthor{jin2017downscaling} used geographically-weighted area-to-area regression Kriging to estimate soil moisture mapping of the upper reaches of Heihe Basin River \cite{jin2017downscaling}.  \citeauthor{su2015prediction} combined regression Kriging with high-resolution multi-spectral imagery to obtain detailed bathymetric mappings of the south shore of Molokai Island, Hawaii \cite{su2015prediction}. In many instances, the solution contains sharp gradients, for example when submerged objects are present. In that case, a Gaussian prior is not adequate. While different choices of covariance models lead to different predictions, none of those predictions preserve patterns with sharp gradients, unless dense observations are available. 

Total variation prior (TV) is another choice for edge-preserving Bayesian inversion problems. \citeauthor{lee2013bayesian} have shown that using a total variation prior can be useful for discrete geologic structure identification \cite{lee2013bayesian}. In particular, it allows for solutions with zonal structures that have only a few large changes at boundary, but small or negligible variations within each zones. However, it has been shown that such prior is not suitable for edge preserving solutions with high resolution, as it converges weakly to a standard Gaussian Process when the discretization of our data grid becomes finer and finer \cite{lassas2004can}. 

Instead of Kriging that requires handcrafted prior models, Deep Neural Networks (DNN) have been used as a data-driven approach to automatically learn highly complex and non-linear mappings, and to provide solutions that have features with sharp gradients. \citeauthor{moses2013lake} developed lake bathymetry for Akkulam–Veli Lake, Kerala, India via the Indian Remote Sensing Imagery and DNN \cite{moses2013lake}. \citeauthor{sharawy2016assessment} estimated shallow water depth in El Burullus Lake by using DNN and high resolution satellite imagery \cite{sharawy2016assessment}. \citeauthor{ceyhun2010remote} used DNN with remote sensing images and sample depth measurements to predict nearshore bathymetry in Fo\c ca's bay, Izmir, Turkey \cite{ceyhun2010remote}. However, unlike Kriging, the applications of DNN in those work do not provide uncertainty quantification, which can be crucial in many engineering applications.
\section{Contribution}

In this work, we propose the use of deep learning techniques within a Bayesian framework that provides uncertainty quantification. We apply it to the interpolation problem of predicting nearshore bathymetry, given sparse point-wise measurements and grid cell average measurements. Our main contributions are as follows:
\begin{itemize}
    \item We propose two deep learning-based approaches that can learn highly non-linear and complex distributions automatically through its training data.
    \begin{itemize}
        \item We trained a conditional Generative Adversarial Network (cGAN) to learn the posterior distribution of nearshore bathymetry. This allows us to sample directly from the posterior distribution and compute different posterior estimates including mean and standard deviation.
        \item We trained a fully connected DNN to estimate the posterior mean. Uncertainty quantification is provided by combining our DNN model with bootstrapping in Kriging. This approach is more computationally feasible than cGAN in terms of both training time and optimization of hyper-parameters.
    \end{itemize}
    \item We propose a method (DNN-Kriging) that uses Kriging to reduce the error in the DNN's prediction of the posterior mean. The motivation for this approach is that the DNN is capable of learning fine-scale features, while Kriging can accurately capture smooth components in the error. Here by ``fine-scale" features we mean features that usually involve rapid oscillations or large values in the derivatives.
\end{itemize}
We compare our methods with Kriging and an interpolation that uses a total variation prior (TV). The rest of the paper is organized as follows. Section \ref{sec:bi} briefly introduces the Bayesian framework, \ref{sec:gan} discusses how conditional Generative Adversarial Networks (cGAN) samples from the posterior distribution, and \ref{sec:dnn} describes how Deep Neural Networks (DNN) computes the posterior mean and generates posterior samples implicitly. Section \ref{sec:dnnkriging} proposes DNN-Kriging. Section \ref{sec:ne} shows numerical benchmarks for Kriging, DNN, cGAN, TV and DNN-Kriging, on smooth solutions as well as solutions with sharp gradients. Section \ref{sec:rw} discusses related work on deep learning-based methods using a Bayesian framework as well as methods that combine DNN and Kriging.



\section{Methodology} 
\subsection{Bayesian Inference} \label{sec:bi}

In this paper, we consider the problem of finding the unknown data $\x$ of the following inverse problem:
\begin{equation}
    \y = H\x + \bv
    \label{eq:meas}   
\end{equation}
where $\x \in \mathbb{R}^n$ is our data, $\y \in \mathbb{R}^m$ is our measurement, $\bv \in \mathbb{R}^m$ is the noise in our measurements and $H$ is a linear forward map. In the framework of Bayesian inference, $\x$, $\y$, and $\bv$ are considered as realizations from random variables $\X$, $\Y$, and $\V$. The unknown data $\X$ is considered to have certain prior distribution $p_{\X}^{\text{prior}}(\x)$, representing what we know about the unknown data before the measurements. The noise $\bv$ follows the distribution $p_{\V}(\bv)$. Given the measurement $\y$, we can then ``update'' the probabilities of the unknown $\X$ from the ``prior'' distribution to a ``posterior'' distribution using the Bayes formula \cite{bolstad2016introduction}:
\begin{equation}
    p^{\text{post}}_{\X}(\x) \propto p_{\V}(\y-H\x) \; p^{\text{prior}}_{\X}(\x)  \label{eq:bayes}
\end{equation}
The posterior distribution $p^{\text{post}}_{\X}$ not only allows us to find the ``most likely'' estimate $\hat{x}$ in the sense that it maximizes the posterior likelihood, but also provides uncertainty quantification that characterizes how confident we are about our prediction. Leveraging deep learning-based methods' ability to learn highly complex and non-linear mappings, we propose using conditional Generative Adversarial Networks (cGAN) to sample from $p^{\text{post}}_{\X}$. We also propose using DNN to estimate the posterior mean $\mathbb{E}(\X | \Y = \y) $, which can be used to generate posterior samples as well through bootstrapping.

\subsection{Generative Adversarial Networks (GAN)} \label{sec:gan}

Generative Adversarial Networks are a class of deep generative models that consists of two types of neural networks \cite{goodfellow2014generative}: a generator $G$ and a discriminator $D$. The generator $G$ takes some noise vector $\z$, which is usually from some normal distribution $p_{\Z}(\z)$, as input to generate fake samples, and the discriminator takes samples as input and tries to classify them as ``real'' or ``fake''. ``Real'' means that the sample is indeed from the target distribution, and vice versa. The networks are trained in an adversarial manner: the generator $G$ tries to generate as realistic samples as possible to fool the discriminator $D$, while $D$ tries to accurately distinguish between ``real'' samples and ``fake'' samples generated by $G$. Formally, let $\x$ represent the bathymetric samples with some prior distribution $p_{\X}(\x)$, then the objective function of GAN will be:
\begin{equation}
\begin{split}
   \min_G \max_D \ &\mathbb{E}_{\x \thicksim p_{\X}(\x)}(\log D(\x))\\
   & + \mathbb{E}_{\z \thicksim p_{\Z}(\z)}(1- \log D(G(\z))) \label{eq:GAN}
\end{split}
\end{equation}
Conditional Generative Adversarial Networks (cGANs) are an extension of GANs \cite{mirza2014conditional}, in which an extra label $\y$, which represents indirect observations, is passed as input to both the generator $G$ and the discriminator $D$. The two networks are trained alternatively using the output of each other, and upon convergence the generator will 
generate samples consistent with observations. 
In this work, two different data types are used for labels $\y$ in our cGAN: 1) point-wise sparse measurements and 2) averages over each grid cell. The objective function of cGAN then becomes:
\begin{equation}
\begin{split}
    \min_G \max_D \ &\mathbb{E}_{\x \thicksim p_{\X}(\x)}(\log D(\x |\y)) \\ 
    & + \mathbb{E}_{\z \thicksim p_{\Z}(\z)}(1- \log D(G(\z|\y))) \label{eq:cGAN}
\end{split}
\end{equation}
Despite GAN's power to directly learn the posterior distribution, it is known that the training process of GAN is, in general, very challenging and requires a significant amount of efforts to tune hyper-parameters.
\subsection{Deep Neural Networks (DNN)} \label{sec:dnn}
Instead of using cGAN, we also consider combining DNN with bootstrapping to sample from the posterior distribution. Our DNN approach relies on the following key mathematical fact. Let $\tau^{*}(\y) =  \mathbb{E}(\X | \Y = \y)$, then $\tau^{*}$ is the solution to the following minimization problem:
\begin{equation}
    \tau^{*} \in \argmin_{\tau:\,\mathbb{R}^m \rightarrow \mathbb{R}^n}{\mathbb{E}(\Vert \X - \tau(\Y) \Vert^2)} 
    \label{eq:mseminimize}
\end{equation}
where the minimization is over all measurable functions from $\mathbb{R}^{m}$ to $\mathbb{R}^{n}$. Empirically, with training data $(\y_i, \x_i)$ that are pairs of realizations of random variable $(\Y, \X)$, we may replace the expectation with the average over training data and solve the following minimization problem:
\begin{equation}
    \min_{\tau:\,\mathbb{R}^m \rightarrow \mathbb{R}^n}{\frac{1}{N}\sum_i^N \Vert \x_i - \tau(\y_i) \Vert^2} 
    \label{eq:avgminimize}
\end{equation}
Given DNNs' ability to approximate highly complex functions, it is then natural to optimize over a family of neural networks with trainable parameter $\theta$:
\begin{equation}
    \min_{\theta}{\frac{1}{N}\sum_i^N \Vert \x_i - \tau_{\theta}(\y_i) \Vert^2} \label{eq:nnminimize}
\end{equation}
Consequently, assuming a high capacity for our DNN and sufficient training data, the optimal neural network $\tau_{\theta^{*}}$ is an estimator for the posterior mean:
\begin{equation}
    \tau_{\theta^{*}}(\y) \approx \mathbb{E}(\X | \Y = \y) 
    \label{eq:nneqposmean}
\end{equation}

We also combine DNN's posterior mean estimator $\tau_{\theta^{*}}$ with bootstrapping from Kriging to generate conditional realizations. To provide motivation, we first briefly review the use of Kriging (or linear Gaussian regression in our case), which assumes a Gaussian prior with mean $\bmu$ and covariance $Q$ for $\X$, as well as a Gaussian noise with zero mean and covariance $R$ for $\V$:
\begin{equation}
    \X \sim \mathcal{N}(\bmu, Q), \;
    \V \sim \mathcal{N}(0, R)
\end{equation}
This implies that our posterior is also a Gaussian distribution for a linear map $H$:
\begin{equation}
\begin{split}
    p^{\text{post}}_{\X}(x) \propto \exp \Big(-\frac{1}{2}(\y-H\x)^TR^{-1}(\y-H\x) \\
    -\frac{1}{2}(\x - \bmu)^T Q^{-1} (\x - \bmu) \Big)   
\end{split}
\end{equation}
Since the forward map $H$ is linear, we can derive the posterior mean in closed form:
\begin{equation}
    \bmu_{\x | \y} = \mathbb{E}(\X | \Y = \y) = \bmu + \Lambda(\y-H\bmu) 
    \label{eq:LGPM}
\end{equation}
where
\begin{equation}
    \Lambda = (Q^{-1} + H^T R^{-1} H)^{-1} H^T R^{-1}
    \label{eq:lbda}
\end{equation}
Uncertainty quantification is represented by the posterior covariance matrix:
\begin{equation}
    Q_{\x | \y} = (Q^{-1} + H^T R^{-1}H)^{-1}    
    \label{eq:postuncert}
\end{equation}
The quantity $\bmu_{\x | \y}$ and $Q_{\x | \y}$ completely characterize the Gaussian distribution $p^{\text{post}}_{\X}(\x)$.  To obtain uncertainty quantification, we can perform a Cholesky factorization $\Lambda = LL^T$, sample $u_i$ from standard Gaussian distribution $\mathcal{N}(0,I)$ and compute conditional realizations $\x_{ci} = Lu_i$. However, sometimes it may be computationally expensive to perform Cholesky decomposition when $n$ is large. Instead, it is easier to generate conditional realizations $\x_{ci}$ implicitly by bootstrapping \cite{kitanidis1995quasi}:
\begin{equation}
    \x_{ci} =\x_{ui} + \Lambda(\y + \bv_i - H\x_{ui}) 
    \label{eq:ui}
\end{equation}
where $\x_{ui}$ is an unconditional realization of $\X$, and $\bv_i$ is a realization of $\bv$. Eq.~\eqref{eq:ui} looks very similar to Eq.~\eqref{eq:LGPM}, in the sense that $\x_{ui}$ can be viewed as a perturbed prior mean, and $\y + \bv_i$ can be viewed as a perturbed measurement. 

We note here that Eq.~\eqref{eq:ui} can also be written as follows using Eq.~\eqref{eq:LGPM}:
\begin{equation}
\begin{split}
    \x_{ci} &=\x_{ui} - \bmu + \bmu +  \Lambda(\y + \bv_i - H(\x_{ui}-\bmu) - H\bmu)  \\ &=\x_{ui} - \bmu + \mathbb{E}(\X | \Y = \y + \bv_i - H(\x_{ui}-\bmu)) \label{eq:generateci}
\end{split}
\end{equation}
With our optimal DNN $\tau_{\theta^{*}}(\y)$ as an estimator for the posterior mean $\mathbb{E}(\X | \Y = \y)$, we can then generate conditional realizations as follows:
\begin{equation}
    \x_{ci}=\x_{ui} - \bmu + \tau_{\theta^{*}}(\y + \bv_i - H(\x_{ui}-\bmu)) \label{eq:dnnci}
\end{equation}
The main advantage of this approach is its computational feasibility. A simple fully connected neural network with two hidden layers can be trained reasonably well to directly estimate the posterior mean. Hence, the total time required for training will be much less than that of GAN. It provides an effective confidence interval with minimal effort of tuning the hyper-parameters and network structures, which is also a significant cost for training GAN.

\subsection{DNN-Kriging} \label{sec:dnnkriging}

\begin{figure}[htbp]
\begin{center}
\includegraphics[width=0.8 \linewidth]{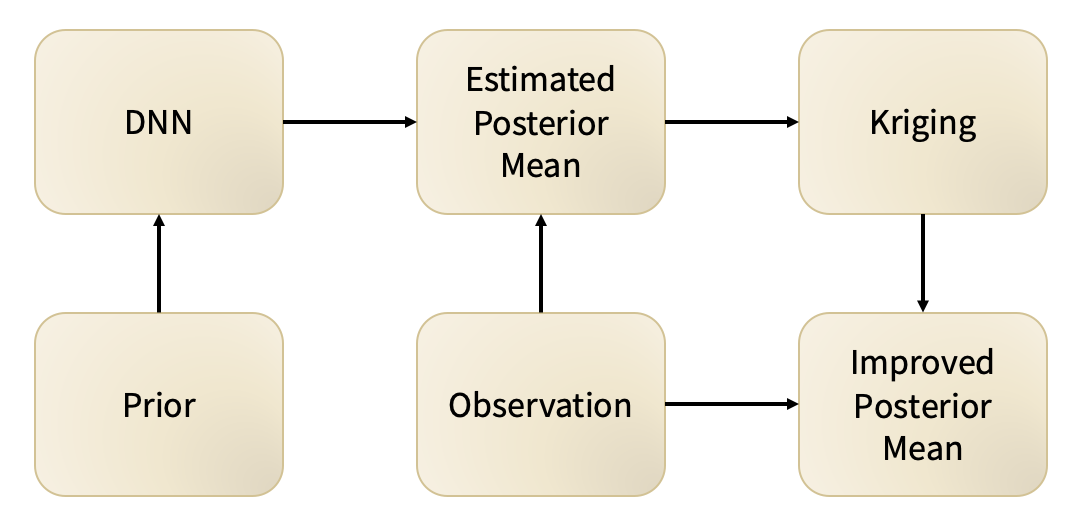}
\caption{DNN-Kriging}
\label{fig:krigingdnn}
\end{center}
\end{figure}

We also propose a method called DNN-Kriging, which is shown in Fig.~\ref{fig:krigingdnn}. The algorithm is a two-step process. In the first step, we use the training data to train our DNN model and make a prediction $\tau_{\theta^{*}}(\y)$ that preserves sharp gradients. Next we replace $\mu$ with $\tau_{\theta^{*}}(\y)$ in Eq.~\eqref{eq:LGPM} to make our final prediction: 
\begin{equation}
    \x_\text{DNN-Kriging} = \tau_{\theta^{*}}(\y) + \Lambda(\y - H \tau_{\theta^{*}}(\y)) \label{eq:dnnKriging}
\end{equation}
where $\Lambda$ is defined in Eq.~\eqref{eq:lbda}. We can generate conditional realizations through the posterior covariance matrix $Q_{\x | \y}$. The motivation for this is that DNN is capable of preserving fine-scale features, while Kriging excels at capturing smooth variations of data. Hence, assuming that the error between the reference profile $\x$ and $\tau_{\theta^{*}}(\y)$ can be characterized by Gaussian priors, we use Kriging to improve DNN's estimates. Our results show DNN-Kriging can provide better estimates than DNN or Kriging used separately, as it combines the advantage of both methods.

\section{Related Work} \label{sec:rw}
\subsection{Deep Learning with Uncertainty Quantification}
Similar deep learning-based methods from a Bayesian perspective have also been proposed by \citeauthor{adler2018deep} to model the posterior distribution of 3D computer tomography images. However, \citeauthor{adler2018deep} did not consider combining DNN with  bootstrapping from Kriging to provide uncertainty quantification. Instead, to predict point-wise posterior standard deviation, \citeauthor{adler2018deep} trained a second neural network $\tau_{\psi^*}$ that minimizes $\mathbb{E}(\Vert \Vert \X - \tau_{\theta^*}(\Y)\Vert^{2} - \tau_{\psi}(\Y) \Vert^2)$. The accuracy of this approach depends very much on the accuracy of the posterior mean estimator $\tau_{\theta^{*}}(\y).$
\citeauthor{patel2019bayesian} used GAN to learn the data distribution and performed Gaussian Process Regression on the noise vector $\z$. Their approach differs from ours as they are using GANs to learn the prior distribution instead of using cGANs to learn the posterior \cite{patel2019bayesian}.

Other attempts to provide uncertainty prediction in DNNs include training an ensemble of neural networks, introducing randomness in network parameters and using Monte-Carlo Dropout \cite{lakshminarayanan2017simple,kononenko1989bayesian,osband2016risk}. However, they often fail to provide posterior distribution, unlike our approaches that apply deep learning from a Bayesian perspective.
\subsection{Combining DNN and Kriging}

Most previous works on combining methods of Kriging and DNNs were established based on the idea of Neural Network Residual Kriging (NNRK), which is a hybrid model that combines DNN with Kriging-based methods. In NNRK, DNN is used to estimate the general spatial trend of data, whereas Kriging is used to estimate the residuals of the DNN's estimate. It has been applied to study a wide range of environmental data and shows improvement over DNN and Kriging \cite{demyanov1998neural,cellura2008wind,padarian2012modelling,dai2014spatial,seo2015estimating}.

\citeauthor{demyanov1998neural} considered direct applications of NNRK for predictions of climate data on Switzerland \cite{demyanov1998neural}. \citeauthor{cellura2008wind} obtained the spatial estimation of the wind field in Sicily via neural Kriging \cite{cellura2008wind}. \citeauthor{padarian2012modelling} used NNRK to model the spatial distribution of soil organic carbon in Chile \cite{padarian2012modelling}. \citeauthor{dai2014spatial} applied NNRK to improve the accuracy of soil organic mapping in Tibetan Plateau at large scale \cite{dai2014spatial}. \citeauthor{seo2015estimating} used regression Kriging to generate training samples and then used NNRK to estimate spatial precipitation in South Korea \cite{seo2015estimating}. Our method DNN-Kriging is an extension of NNRK to a general Bayesian formulation.

\section{Numerical Experiments} \label{sec:ne}

In this section, DNN will refer to a fully connected neural network with 2 hidden layers and 2{,}000 neurons in each layer and is used as a posterior mean estimator $\tau_{\theta^{*}}$\footnote{Detailed implementation of DNN and cGAN can be found at https://github.com/qyz96/DeepBayesian}. The weights are initialized at random with a uniform distribution, and the Adam optimizer is used in training. We consider an interpolation problem of nearshore bathymetry on a rectangular domain uniformly discretized with 51 $\times$ 75 grid points, at Field Research Facility (FRF) in Duck, North Carolina. Monthly surveys with units in meters are provided by the U.S.\ Army Corps of Engineer Field Research Facility. Our measurements are multi-scale: they include point-wise sparse measurements (i.e., direct survey) as well as grid cell averages (e.g., remote sensing survey). 239 direct bathymetric surveys are used to generate synthetic training data by adding Gaussian noise with covariance matrix $C$ such that $C_{ij}=\alpha\exp(-d^2/r^2)$, where
$$d=\sqrt{\left(\frac{x_i-x_j}{L} \right)^2 + \left(\frac{y_i-y_j}{W} \right)^2}$$
$(x_i, y_i)$, $(x_j, y_j)$ are the coordinates of grid points $(i, j)$, $r = 0.07$, $\alpha = 0.15$, $L$ and $W$ are the length and width of the field respectively. In our synthetic training data, we also add ``random'' rectangular jumps representing sandbars and other discrete nearshore features with the following strategy. A rectangle region is chosen with fixed size ($0.44 \, W$ by $0.44 \, L$), the coordinate of the top left corner is uniformly sampled from $[0,0]$ to $[W,L]$, and a constant value of 12 m is added to that rectangular region. 95,600 samples are generated by adding Gaussian noise to real bathymetric surveys for data augmentation as well as describing smooth variations in bathymetry (we generate 400 synthetic training samples based on each survey); for 47,800 of them we add the random rectangular jumps. 
We also add Gaussian white noise to all training inputs with different scales for different data types. The variance of white noises for point-wise and grid cell average measurements are 0.01m and 0.001m respectively. In Kriging, $\mu$ is chosen to be the average over those 239 surveys. The covariance matrix is given by $Q = 10^{\theta_1} Q_0$, where $Q_0$ is defined using the exponential kernel (i.e., $[Q_0]_{i,j}=\exp(-d/r)$ with $r=0.75$). The error matrix $R = 10^{\theta_2} R_0$ is defined as follows. Let $\y = [\y_p,\y_g]$, where $\y_p$ and $\y_g$ represent point-wise measurements and grid cell average measurements, respectively. $R_0$ is a diagonal matrix with $[R_0]_{pp} = 0.01 \, I$ and $[R_0]_{gg} = 0.001 \, I$. The hyper-parameters $\theta_1$ and $\theta_2$ are estimated by maximizing the marginal likelihood \cite{kitanidis1983statistical}, which we achieve by sampling a uniform grid in $(\theta_1,\theta_2)$. We use 35 evenly distributed point-wise measurements and 24 grid-cell average measurements.

\subsection{Estimation of Posterior Mean and Uncertainty Quantification}

\begin{figure}[htb]
\begin{center}
\includegraphics[width=1.0 \linewidth]{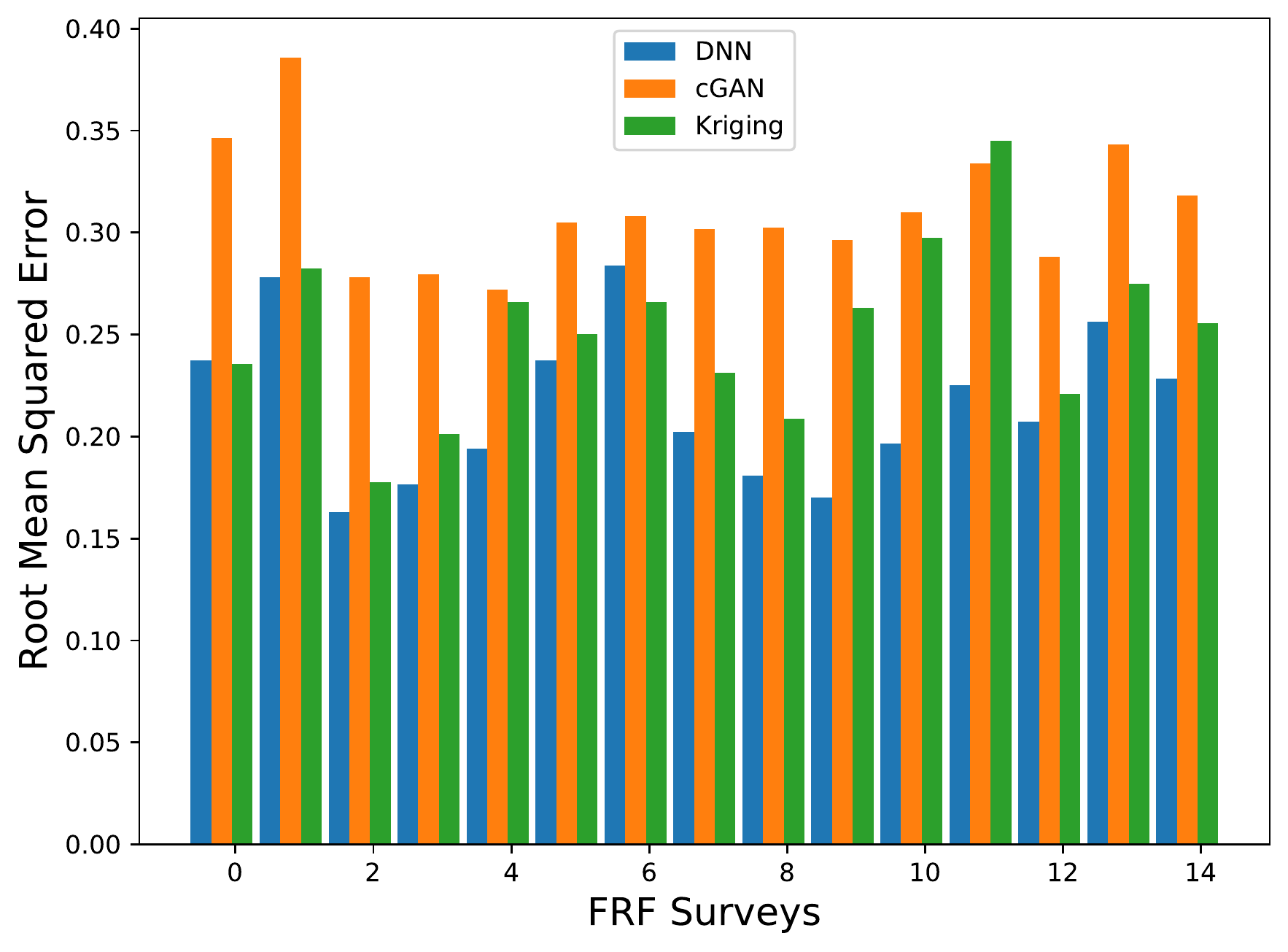}
\caption{Comparison of conditional GAN (cGAN), Deep Neural Networks (DNN), and Kriging on FRF surveys.}
\label{fig:benchmark}
\end{center}
\end{figure}

We first compare cGAN and DNN with Kriging based on 15 real FRF surveys not included in the training set. For each prediction, 1,000 samples are generated by cGAN and we use the point-wise average over those samples as our final estimate. Fig.~\ref{fig:benchmark} shows the root mean squared errors of cGAN and Kriging for those 15 surveys. The average root mean squared errors for DNN, cGAN and Kriging are 0.215, 0.311 and 0.252, respectively. Our results show that Kriging performs better than cGAN in most cases, while DNN's estimates have the lowest root mean squared errors on 13 out of 15 surveys.

\begin{figure}[htbp]
\begin{center}
\includegraphics[width=1.0 \linewidth]{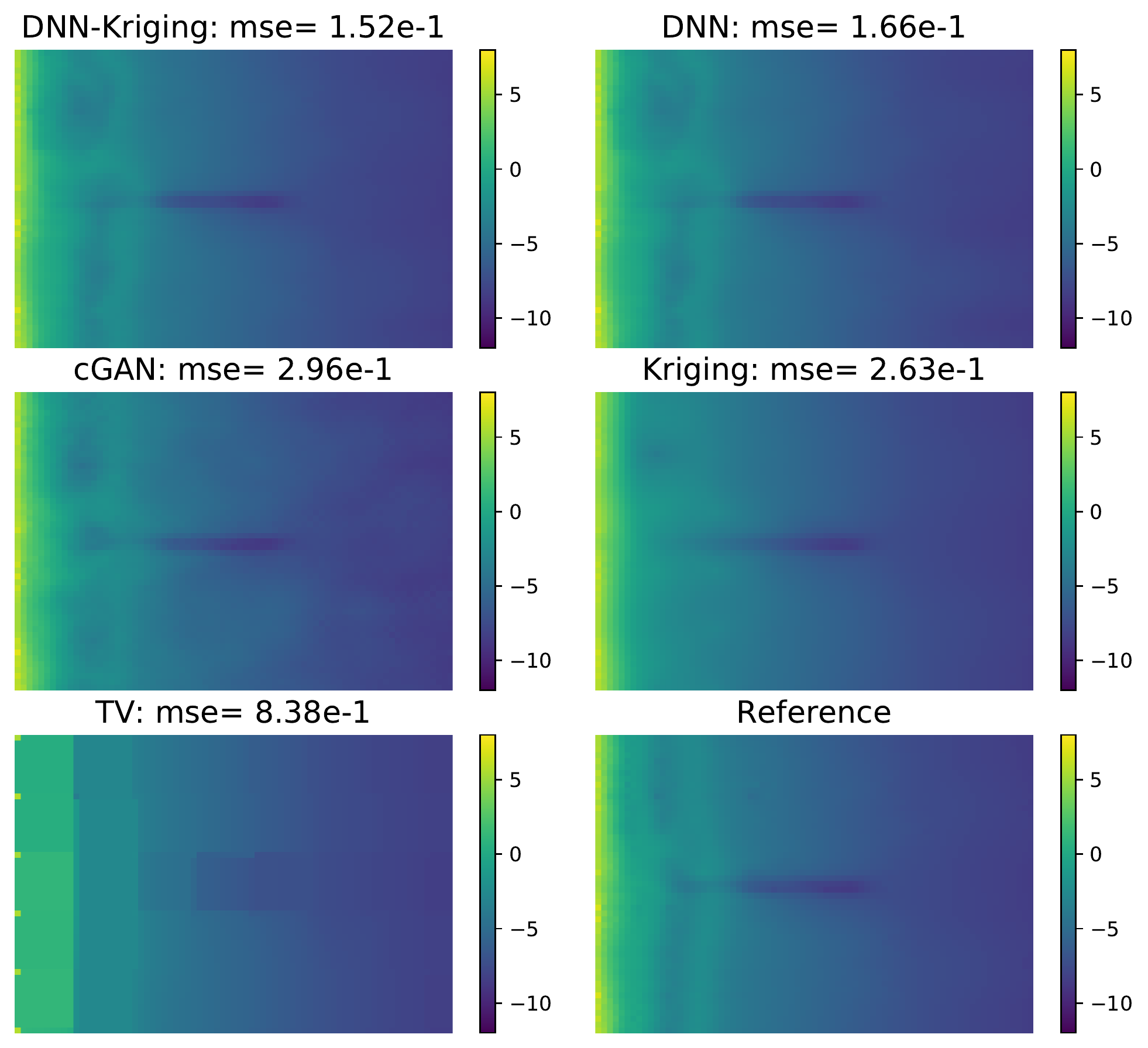}
\caption{Comparison of Kriging, TV (Total Variation), DNN-Kriging, cGAN, and DNN on FRF survey. ``mse'' (mean squared error) is with respect to ``Reference.''}
\label{fig:surveycomp}
\end{center}
\end{figure}

\begin{figure}[htbp]
\begin{center}
\includegraphics[width=1.0 \linewidth]{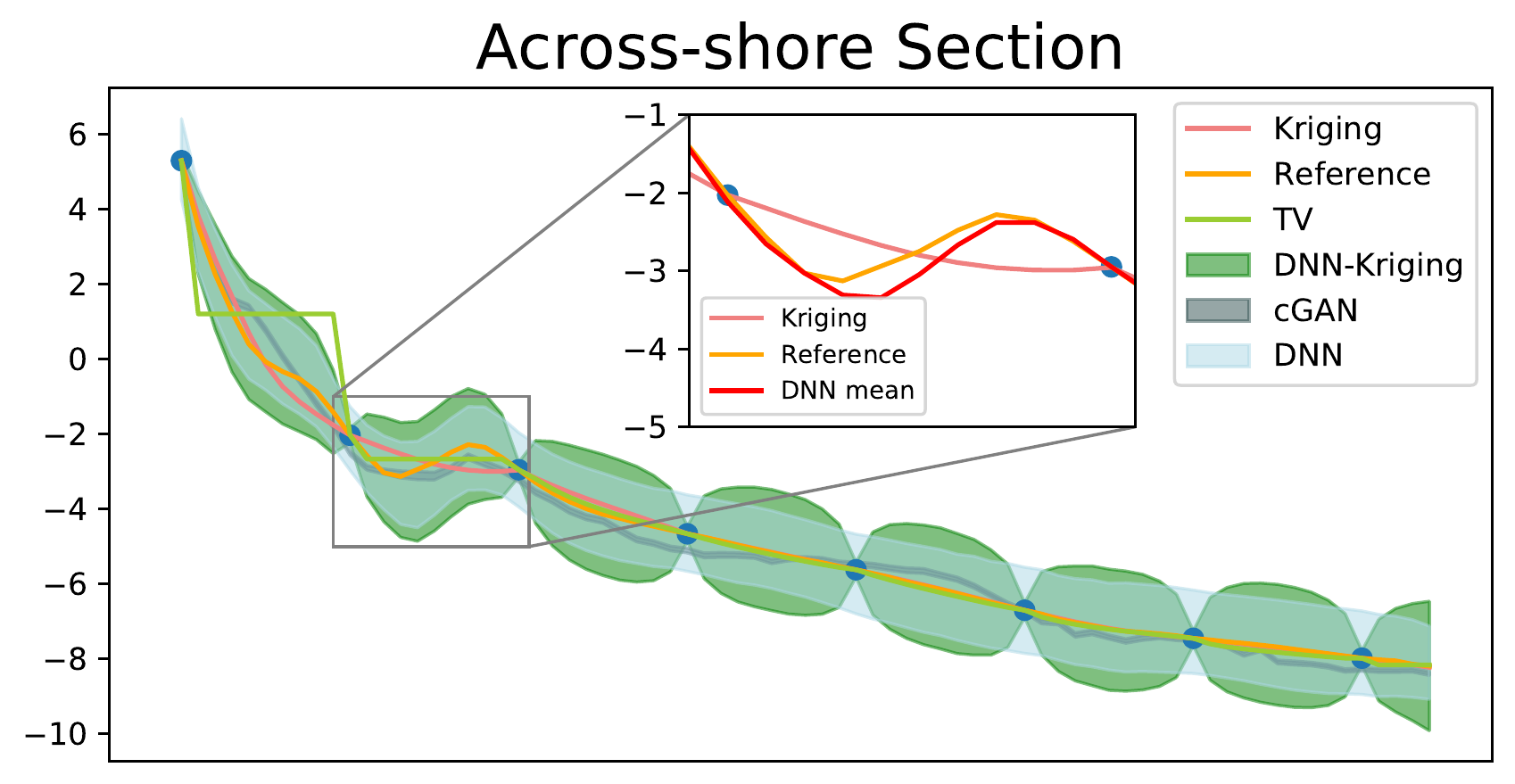}
\caption{Across-shore comparison of Kriging, TV (Total Variation), DNN-Kriging, cGAN, and DNN.}
\label{fig:ass}
\end{center}
\vspace*{-5mm}
\end{figure}

\begin{figure}[htbp]
\begin{center}
\includegraphics[width=1.0 \linewidth]{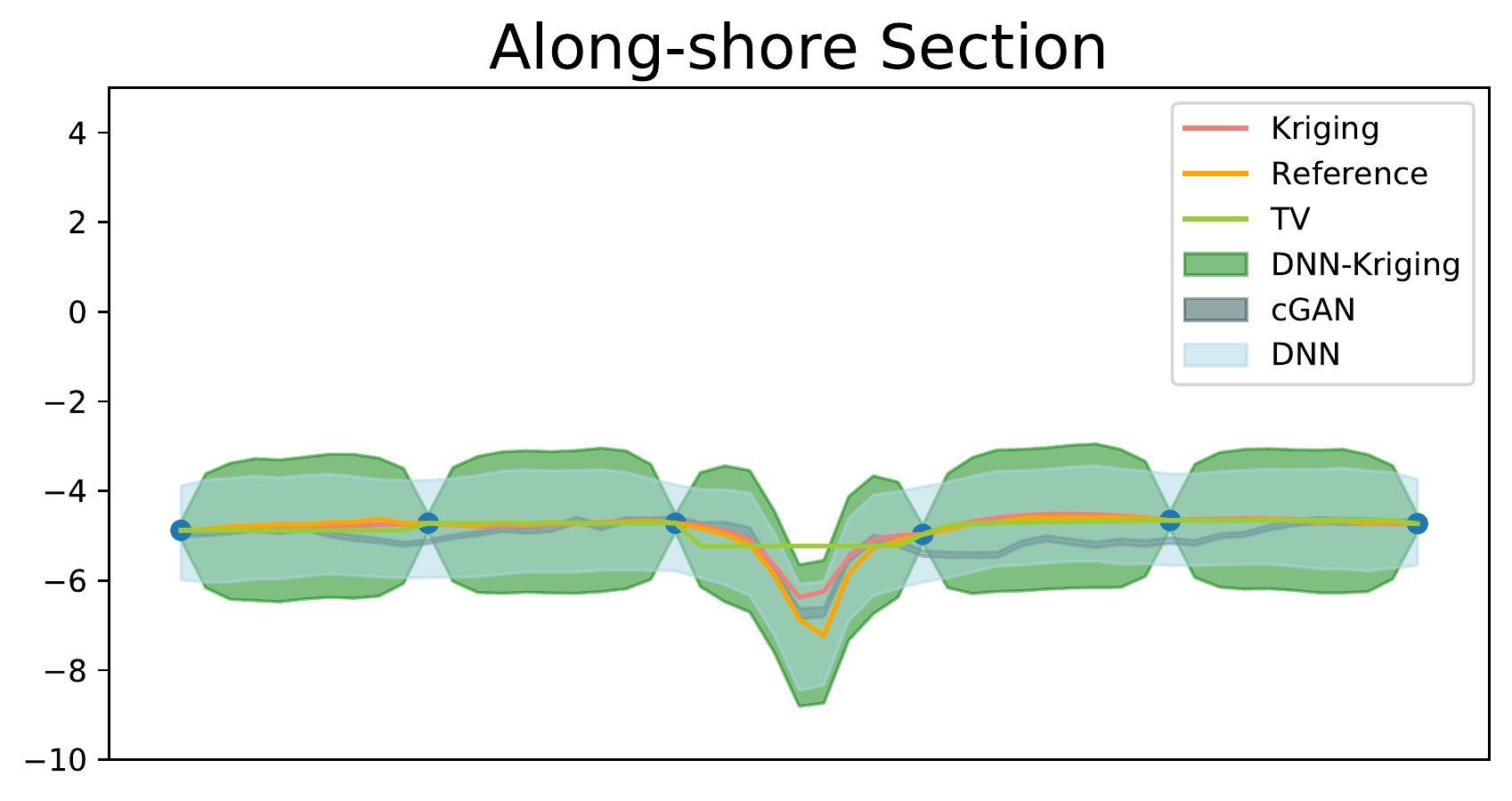}
\caption{Along-shore comparison of Kriging, TV (Total Variation), DNN-Kriging, cGAN and DNN.}
\label{fig:als}
\end{center}
\end{figure}

We compare the performance of DNN-Kriging, DNN, cGAN, Kriging and TV on one FRF survey as shown in Fig.~\ref{fig:surveycomp} with respect to the reference profile. We see that DNN-Kriging gives the best estimate. We also show the along-shore and across-shore section plots of our predictions, along with their Bayesian confidence intervals in Fig.~\ref{fig:ass} and Fig.~\ref{fig:als}. We compute the DNN-Kriging's confidence interval using the posterior covariance matrix from Kriging. The confidence interval of the DNN's prediction is obtained by bootstrapping as shown in Eq.~\eqref{eq:dnnci}. We perform a posterior sampling with cGAN and compute the corresponding confidence intervals. We see that the confidence interval of DNN is smaller than the one produced by Kriging. The uncertainty of cGAN is unreasonably small and does not include the reference profile. This is probably due to the mode collapse \cite{thanh2018catastrophic}. 


In Fig.~\ref{fig:ass}, we also observe that Kriging gives a smoother prediction than deep learning-based methods near the ``bump'' in the middle left side of the plot that represents sandbar structures. This is expected since Kriging assumes a widely used Gaussian (i.e., exponential) prior distribution that promotes smooth variations in the bathymetry.


\begin{figure}[htb]
\begin{center}
\includegraphics[width=1.0 \linewidth]{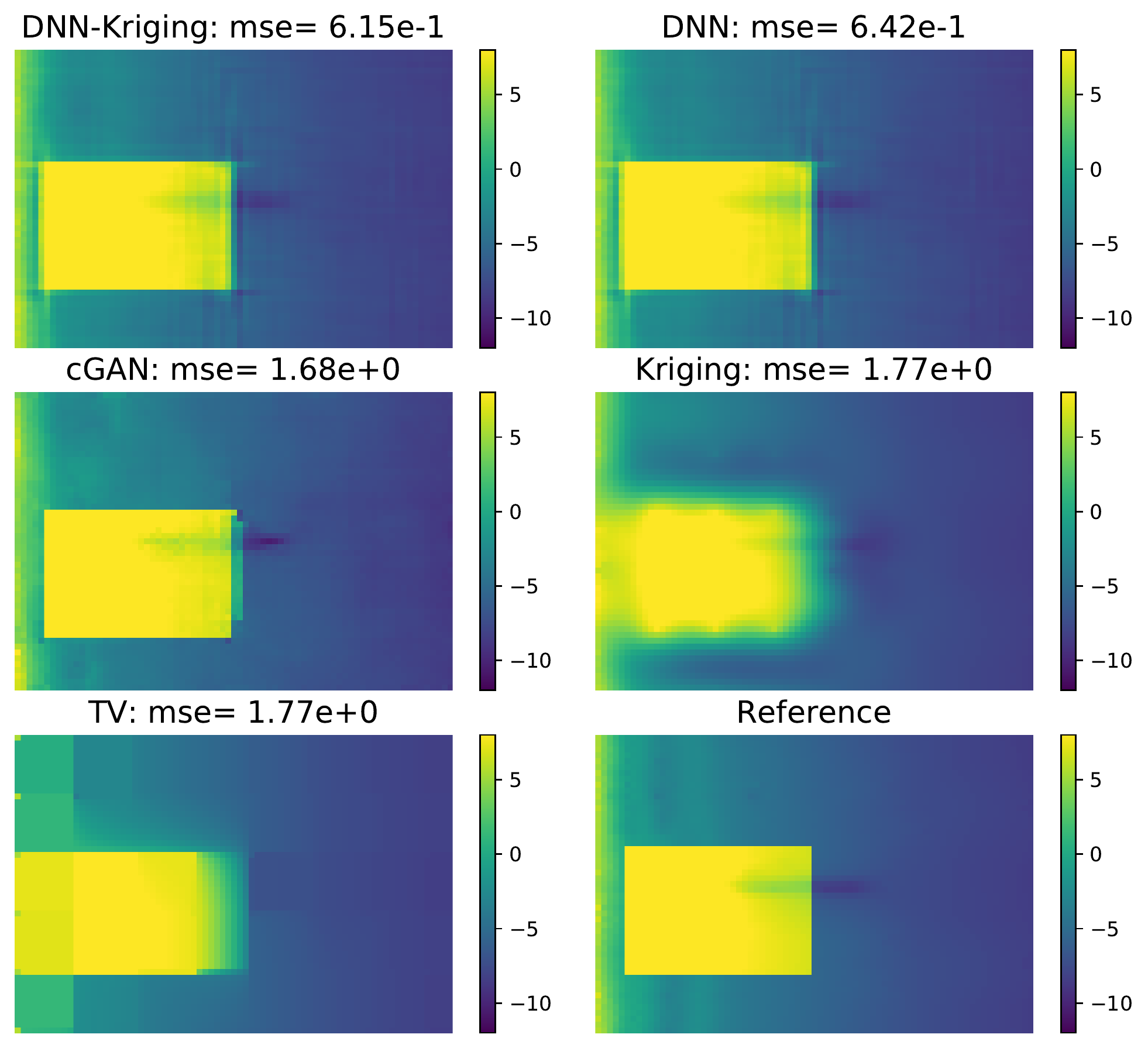}
\caption{Comparison of Kriging, TV (Total Variation), DNN-Kriging, cGAN, and DNN on bathymetric survey with a rectangular ``jump.''}
\label{fig:jbm}
\end{center}
\end{figure}

\begin{figure}[htb]
\begin{center}
\includegraphics[width=1.0 \linewidth]{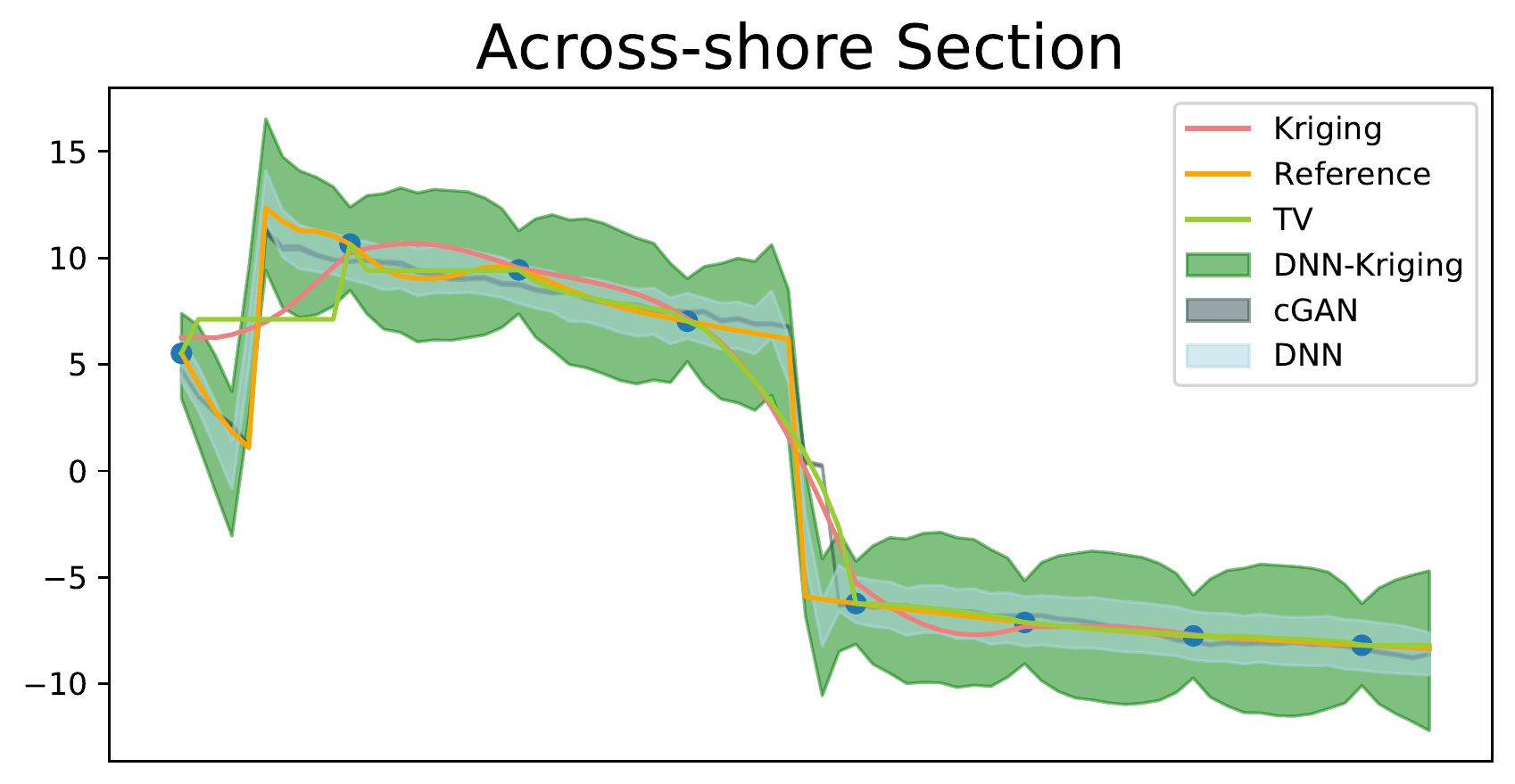}
\caption{Across-shore comparison of Kriging, TV (Total Variation), DNN-Kriging, cGAN, and DNN.}
\label{fig:asj}
\end{center}
\end{figure}

\begin{figure}[htb]
\begin{center}
\includegraphics[width=1.0 \linewidth]{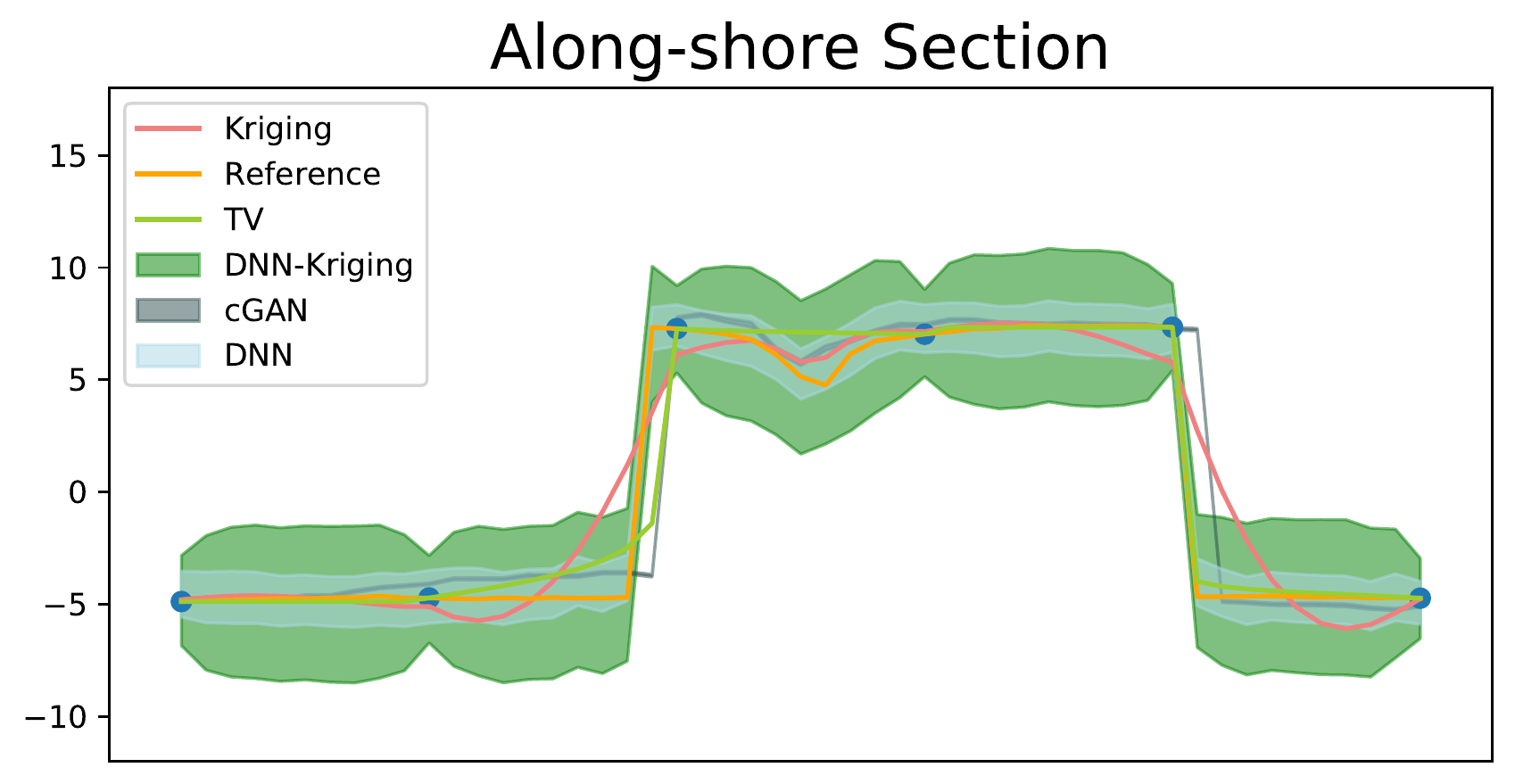}
\caption{Along-shore comparison of Kriging, TV (Total Variation), DNN-Kriging, cGAN, and DNN.}
\label{fig:acj}
\end{center}
\end{figure}

In Fig.~\ref{fig:jbm}, one FRF survey (collected on June 21th, 2017) is chosen and a random jump is added for performance comparisons of cGAN, DNN, Kriging, DNN-Kriging, and TV (Kriging with a total variation prior). Similarly, 1000 samples are generated by cGAN to compute the average as its final prediction. Fig.~\ref{fig:asj} and Fig.~\ref{fig:acj} show two cross section plots selected near the location of the jump with the corresponding confidence intervals. We observe that cGAN, DNN, and DNN-Kriging give estimates with almost vertical jumps, whereas Kriging provides prediction that transitions linearly near the location of the jump. Although Kriging with a total variation prior encourages rapid changes in small intervals, we can still see transitions that are linear or similar to staircase functions near the location of the jump. Here, we emphasize again the important role that prior information has played in our methods. With a smooth Gaussian prior, Kriging gives solutions with linear transitions near the location of the jump that minimize the mean squared error. However, with a training data set that has prior information regarding the presence of rectangular jumps in our bathymetry, our DNN performs a piece-wise regression capable of providing estimates with sharp jumps. We also remark that DNNs learn to express highly complex functions automatically through its training data~\cite{poole2016exponential}, while Kriging requires a carefully chosen nonlinear kernel function~\cite{williams1996computing}.

\begin{figure}[htb]
\begin{center}
\includegraphics[width=1.0 \linewidth]{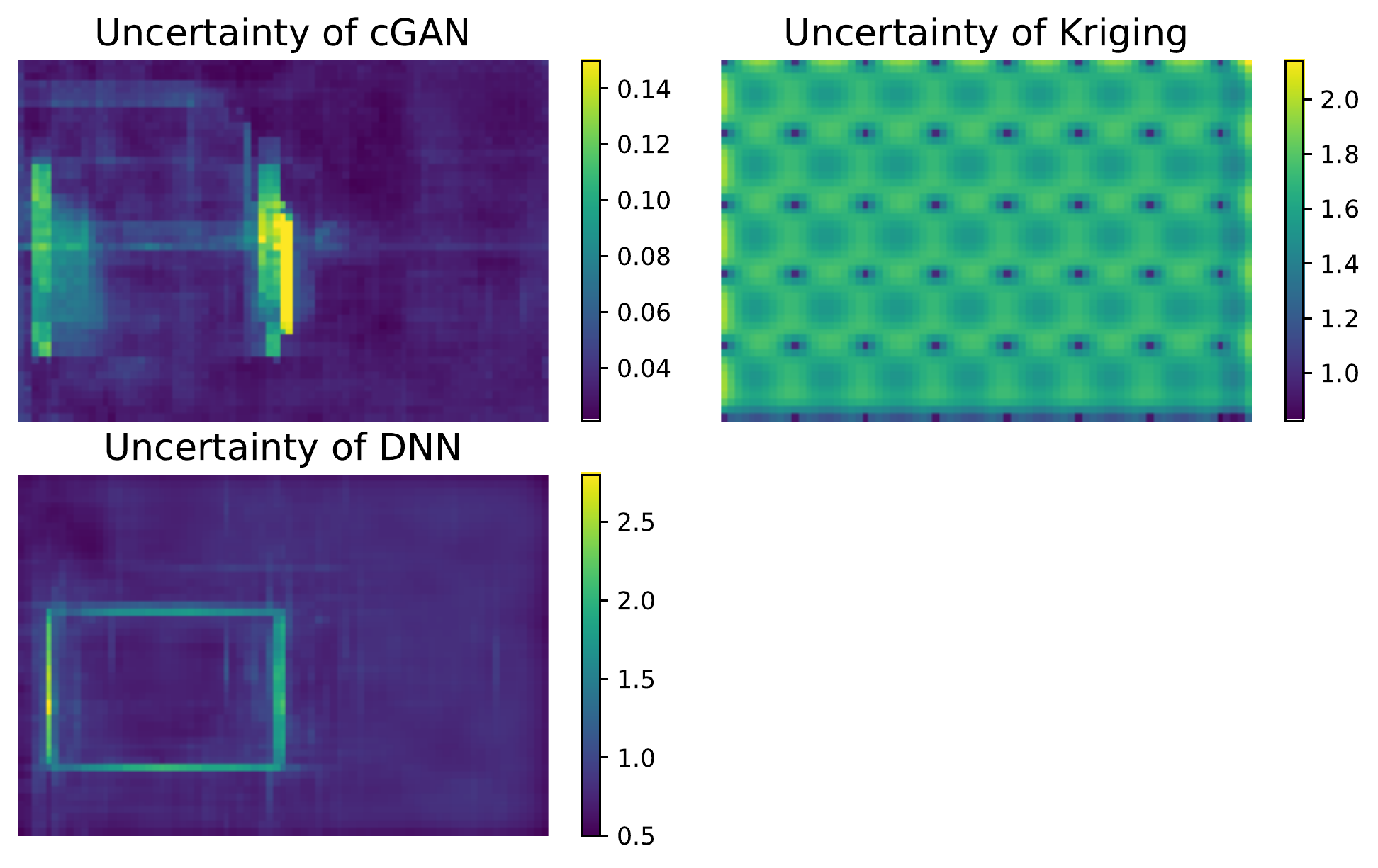}
\caption{Uncertainty Quantification by Kriging, DNN, and cGAN.}
\label{fig:uncertainty}
\end{center}
\end{figure}

In Fig.~\ref{fig:jbm}, we also see that DNN-Kriging provides more accurate estimates than pure DNN models. Intuitively, DNN, which knows about the presence of rectangular jumps from its training data, learns the approximate location of the jump and estimates the posterior mean accordingly. Capable of predicting solutions with smooth variations through its Gaussian prior, Kriging improves the posterior mean estimate by predicting the error between the unknown data $\x$ and the DNN's prediction $\tau_{\theta^{*}}(\y)$. Here, we also illustrate the impact of a different prior in terms of uncertainty quantification. Fig.~\ref{fig:uncertainty} shows the posterior standard deviation provided by Kriging, DNN, and cGAN. We observe here that both DNN and cGAN appear to have relatively high uncertainty near the boundary of the rectangular jump. Due to the presence of noise in measurements and sparse measurement locations, the location of our jump is uncertain, which causes high uncertainty in the corresponding area. In Kriging, the assumption of a Gaussian prior makes its posterior uncertainty \textbf{independent of the  measurements} (see Eq.~\eqref{eq:postuncert}), while the uncertainty prediction by both DNN and cGAN varies based on the measurements.

\section{Conclusion}

In this work, we have explored the use of deep learning techniques from a Bayesian perspective to estimate nearshore bathymetry with sparse point-wise measurements and grid cell average measurements. We proposed using conditional Generative Adversarial Networks (cGAN) as a purely data-driven approach to directly sample the posterior distribution. This usually has a challenging training process and requires extra effort for tuning of hyper-parameters. We also proposed using a fully connected DNN to directly estimate the posterior mean of the bathymetry, which can be combined with bootstrapping from Kriging to provide uncertainty quantification. Both approaches provide more accurate predictions than Kriging when sharp changes are present in nearshore surveys. Finally, we proposed a method named DNN-Kriging that combines Kriging's ability to model smooth variations in the residuals with DNN's ability to capture fine-scale features. Results show that DNN-Kriging provides the best estimate among all the methods.

\section{Acknowledgement}

This research was supported by the U.S. Department of Energy, Office of Advanced Scientific Computing Research under the Collaboratory on Mathematics and Physics-Informed Learning Machines for Multiscale and Multiphysics Problems (PhILMs) project, PhILMS grant DE-SC0019453.


\bibliographystyle{aaai}
\bibliography{AAAI2020}
\end{document}